\documentclass[smallcondensed,natbib]{svjour3} 

\usepackage[table]{xcolor}
\definecolor{color2}{RGB}{33,72,131} % blue
\definecolor{color2b}{RGB}{33,72,131} % blue
\usepackage{tikz}
\newcommand{\pie}[1]{%
	\begin{tikzpicture}
	\color{color2}
	\draw (0,0) circle (1.4ex);\fill (1.4ex,0) arc (0:#1:1.4ex) -- (0,0) -- cycle;
	\end{tikzpicture}%
}
\newcolumntype{C}[1]{>{\centering\let\newline\\\arraybackslash\hspace{0pt}}m{#1}}
\newcommand{\mcrot}[4]{\multicolumn{#1}{#2}{\rlap{\rotatebox{#3}{#4}~}}} 
\newcolumntype{L}[1]{>{\raggedright\let\newline\\\arraybackslash\hspace{0pt}}m{#1}}

\usepackage{graphicx}
\usepackage{hyperref}
\usepackage{booktabs}
\usepackage{pifont}
\usepackage{tabularx}
\usepackage{verbatim}

\newcommand*\rot{\rotatebox{90}}
\newcommand*\ok{\ding{51}}

\journalname{}
\title{A systematic literature review on state-of-the-art deep learning methods for process prediction}
\titlerunning{A systematic literature review on deep learning methods for process prediction}
\author{Dominic A. Neu \and Johannes Lahann \and Peter Fettke}
\authorrunning{Neu, Lahann \& Fettke}
\institute{German Research Center for Artificial Intelligence (DFKI), Saarbruecken, Germany \newline Saarland University, Saarbruecken, Germany  \email{dominic.neu@dfki.de}}

\begin{document}
\maketitle
\begin{abstract}
Process mining enables the reconstruction and evaluation of business processes based on digital traces in IT systems. An increasingly important technique in this context is process prediction. Given a sequence of events of an ongoing trace, process prediction allows forecasting upcoming events or performance measurements. In recent years, multiple process prediction approaches have been proposed, applying different data processing schemes and prediction algorithms. This study focuses on deep learning algorithms since they seem to outperform their machine learning alternatives consistently. Whilst having a common learning algorithm, they use different data preprocessing techniques, implement a variety of network topologies and focus on various goals such as outcome prediction, time prediction or control-flow prediction. Additionally, the set of log-data, evaluation metrics and baselines used by the authors diverge, making the results hard to compare.
This paper attempts to synthesise the advantages and disadvantages of the procedural decisions in these approaches by conducting a systematic literature review.

\end{abstract}
\section{Introduction}
Today's information systems create, utilize and store vast amounts of data about the business processes being executed with them. These logs capture the as-is execution. Process mining extracts knowledge from these logs to provide means for process discovery, process monitoring and process improvement. Additionally, in case target process models are provided, conformance-checking searches for deviations of this model. Consequently, process mining is situated between the disciplines of data mining and business process modelling \citep{Aalst.2011}. In recent years, much effort was put into process discovery to build human-readable models for further investigation by domain experts \citep{Augusto.2018}. These models support descriptive, retrospective and holistic analyses, usually performed by business process managers. However, these models are not able to provide operational decision support at run time.

To fill this gap, a new technique, referred to as predictive business process monitoring or process prediction, has been evolved. This subfield of process mining focuses on single process instances rather than process models and tries to gain predictive insights into the future of this instance. Organizations can leverage this predictive information to adapt the ongoing process execution and thus prevent undesired outcomes \citep{Aalst.2010,Maggi.2014}. As the basis for  predictive models, the usage of machine learning has gained considerable interest. \cite{DiFrancescomarino.2017} identified 62 tools, with the majority applying machine learning or statistical models. The prevalence of these predictive systems may be due to their context agnostic nature, i.e. the same predictive algorithm can be used in multiple business cases with little to none modification. The same would not be valid for rule-based decision support systems.

In the last three years, deep learning-based process prediction approaches gained popularity and led to breakthrough results \citep{Verenich.2019}, e.g. the first publications on this topic \citep{Evermann.2017,Tax.2017} gained more than 100 citations, according to Google Scholar. Deep learning builds upon multiple layers of artificial neurons. Through this hierarchical structure, artificial neural networks do not rely on handcrafted features but are able to learn complex features on their own. This ability reduces the customization need for algorithms to work on a wide range of prediction problems. Also, the learned feature representation is not limited by human imagination but can be arbitrarily complex. Furthermore, the computational requirements and achievable performance scale linearly to the available data, whereas standard process discovery algorithms suffer from quadratic or even exponential run time requirements \citep{Augusto.2018}.

Although deep learning describes one common learning method, the actual implementations used for process prediction are fundamentally diverse. They differ in data preprocessing, implement different network architectures or focus on varying goals such as outcome prediction, time prediction, and control-flow prediction. Additionally, the set of log-data, evaluation metrics and baselines diverges, which makes the results often incomparable. To overcome this issue, the goal of this paper is to give a comprehensive overview of deep learning-based predictive process monitoring approaches and discuss the trade-offs between the existing approaches. In particular, the contribution of this work is threefold:
\begin{enumerate}
\item A systematic literature review (SLR) is conducted to create a comprehensive presentation of existing deep learning-based predictive process monitoring approaches.
\item The identified literature is classified along selected criteria to extract the main contribution of the approaches.
\item Conflicting statements and research gaps are brought to attention in order to generate impulses for further research.
\end{enumerate}

The remainder of this paper is structured as follows. Section 2 introduces the reader to relevant terminology in the field of predictive process monitoring that is required to follow the rest of the paper. Section 3 explains the methodology of the conducted SLR. Afterwards, section 4 presents the investigated literature and further classifies it along multiple dimensions. Section 5 discusses critical and sometimes contradictory findings and provides impulses for further research. Section 6 places the work in the context of the current body of knowledge by examining related work, and Section 7 concludes the paper with a summary and future work.
\section{Related Work}
In the last years, a few surveys tackling the domain of deep learning-based process prediction have been proposed. In this section, we present an overview of the related work and clarify the significant differences between this study and the existing surveys.

\cite{MarquezChamorro.2018} reported a previous survey of process prediction models. The authors analyzed 39 existing process prediction methods and categorized them into process-aware and non-process-aware approaches, i.e. approaches that do or do not exploit an explicit representation of the process model. Also, they distinguished between regression and classification models according to the prediction target of the analyzed approaches. For each paper, they listed the prediction quality, the prediction metric, the name of the data set or the number of traces, the internal prediction algorithm, and the prediction target. The survey does not include any approaches, utilizing deep learning and therefore, does not overlap with this work.

\cite{DiFrancescomarino.2018} developed a value-driven framework for classifying existing process prediction methods to support organizations to navigate in the predictive process monitoring field. This work analyzes 51 approaches with different prediction architectures. The papers are categorized along with their prediction targets in time, categorical outcome, sequence of values, risk, inter-case metrics and costs. Afterwards, they are compared based on input data, tool support, application domain, and algorithm family. The survey covers but does not focus on deep learning-based approaches. In contrast to this study, the survey of \cite{DiFrancescomarino.2018} is only concerned with a qualitative comparison and does not include a quantitative comparison of the analyzed approaches. Furthermore, the survey does not tackle current deep learning methods for process prediction that have led to significant prediction quality increases.

\cite{Teinemaa.2019} conducted a systematic review and created a taxonomy of outcome-oriented predictive process monitoring. In the review, 14 papers were identified and compared according to prefix extraction, filtering, trace bucketing, sequence encoding, and classification algorithm. Furthermore, an experimental evaluation was carried out, testing the impact of different qualitative criteria. For this evaluation, the authors relied on their implementation. The survey of \cite{Teinemaa.2019} does not intersect with this work, since it does not include any deep learning-based process prediction methods. Besides, it only deals with outcome-oriented prediction, whereas this work covers multiple prediction targets.

\cite{Verenich.2019} created a survey on remaining time prediction methods. They identified 25 relevant papers from 2008 to 2017 and compared them based on input data, process awareness, prediction algorithm and application domain. They created quantitative comparability by performing a benchmark of 16 remaining time prediction methods on a selection of publicly available data sets. They covered several instances of LSTMs with different hyperparameter settings. While there are a few overlaps with this study, both papers vary regarding their focuses. Our work investigates the design decisions of neural network architecture and their effects on a more sophisticated level of granularity. In contrast, their work is more interested in the performance differences of process-aware vs non-process-aware methods.

Recently, another survey was published by \cite{Harane.2020}, dealing with deep learning approaches in predictive business process monitoring. However, the authors did not conduct a systematic literature review but compared three existing approaches, namely \cite{Evermann.2016,Mehdiyev.2017,Tax.2017} on a high abstraction level. Our work, on the other hand, covers all deep learning-based approaches according to carefully defined inclusion and exclusion criteria of the literature review and thus captures a much more comprehensive picture of the current state of deep learning methods for process prediction.

\begin{table}[ht]
\caption{Comparison of this study with existing process prediction literature reviews}
\begin{tabular}{  L{6cm} cccccc}
\toprule
 \textbf{Distinguishing factor}   & \multicolumn{6}{c}{\textbf{Reviews}}\\    
	 & \mcrot{1}{l}{65}{\citeauthor{MarquezChamorro.2018}} & \mcrot{1}{l}{65}{\citeauthor{DiFrancescomarino.2018}} & \mcrot{1}{l}{65}{\citeauthor{Teinemaa.2019}} & \mcrot{1}{l}{65}{\citeauthor{Verenich.2019}} & \mcrot{1}{l}{65}{\citeauthor{Harane.2020}} & 
	 \mcrot{1}{l}{65}{\textbf{This Study}} \\
	\midrule
    Release year & 2017 &2018 &2019&2019 &2020 &2020\\
	Total number of papers reviewed & 39 &51 &14&25 &3 &32\\
	Papers with Deep Learning focus & 1 &8 &0&3 &3 &32\\

	\midrule
	Qualitative performance comparison of existing process prediction approaches &
	\pie{200} &\pie{180} & \pie{120}&\pie{0} &\pie{120} &\pie{360}\\
	Detailed definition of inclusion and exclusion criteria &
	\pie{160} &\pie{210}&\pie{360} &\pie{360} &\pie{0} &\pie{360}\\
	Consideration of multiple prediction types and grouping of existing approaches &
	\pie{360} &\pie{200} & \pie{0}&\pie{0} &\pie{120} &\pie{360}\\
	\textit{Focus on the characteristics of Deep Learning and the existing Deep Learning architectures} &
	\pie{10} &\pie{60} & \pie{0}&\pie{50} &\pie{240} &\pie{360}\\
	Consideration of further influencing factors such as pre-processing and  data splitting &
	\pie{220} &\pie{0} & \pie{180}&\pie{280} &\pie{180} &\pie{360}\\
	\midrule
	Quantitative performance comparison of existing process prediction approaches &
	\pie{240} &\pie{0} & \pie{180}&\pie{180} &\pie{180} &\pie{360}\\
	Assessment of the performance score of multiple prediction types &
	\pie{360} &\pie{0} & \pie{0}&\pie{0} &\pie{0} &\pie{360}\\
	Coverage and comparison of a variety of process prediction datasets &
	\pie{120} &\pie{0} & \pie{240}&\pie{360} &\pie{200} &\pie{360}\\
	
	\bottomrule
\end{tabular}
\label{fig:comparision_related work}
\end{table}

Table \ref{fig:comparision_related work} summarizes the central distinguishing factors between this review and existing related work. The table clearly shows that each related work has a different focus.  The most similar work to this study is the review of \citeauthor{MarquezChamorro.2018}. However, it was performed before the rise of deep learning and therefore does not cover deep learning based process prediction.  The major objective of the work by \citeauthor{DiFrancescomarino.2018} is to develop a framework that supports practitioners in selecting the right process prediction algorithm for a specific use case. It therefore only evaluates qualitative criteria and leaves out a quantitative comparison. In addition, the work does not focus on deep learning methods but only includes a few deep learning approaches. Both, the works of \citeauthor{Teinemaa.2019} and \citeauthor{Verenich.2019} focus on one specific prediction type. The former is concerned with outcome-oriented prediction and the latter deals with remaining time prediction. Both papers do not focus on deep learning. In fact, the benchmark performed in \citeauthor{Teinemaa.2019} does not include any deep learning approach at all and \citeauthor{Verenich.2019} reviews only a few deep learning approaches. The work of \citeauthor{Harane.2020} presents only three existing deep learning based approaches. Furthermore, the methodology used to select these papers is not presented in an understandable and reproducible manner.
In contrast to the related work, this study presents the first structured and comprehensive literature review on deep learning-based process prediction. The major objective of this study is to collect and compare existing deep learning approaches for process prediction covering multiple process prediction types such as next-step prediction, outcome prediction, remaining time prediction, etc.

\section{Background}

\subsection{Deep learning}

Deep learning refers to multi-layer neural networks as opposed to shallow machine learning (decision trees, support vector machines). Recent advances in learning algorithms and computational performance made deep learning feasible for many complex prediction tasks. Especially in natural language processing and image classification, deep neural networks have shown superior performance over their machine learning alternatives \citep{LeCun.2015}. In this section, we will provide a brief introduction to the basic concepts, relevant for this review. \citeauthor{ LeCun.2015} gives a more comprehensive overview. For further information on the implementation details, please refer to their references.

The most straightforward neural network architecture is the feed-forward multi-layer perceptron. Several horizontally aligned neurons constitute a layer, all of which connect to neurons of the previous and the following layer. Each neuron calculates weighted arithmetic mean of all the outgoing signals of these previous neurons. An activation function is applied to this single scalar to achieve non-linearity, and the signal propagates to the next layer of neurons. The weights of the arithmetic mean are the model parameters. They are iteratively trained by back propagating the error between the prediction of the network and the correct result.    

Convolutional neural networks are distinct architectures, initially developed for images. A convolutional layer uses a fixed size kernel to swipe over the input. The result is a matrix of many multiplications of the fixed kernel and local areas of the image. This approach was developed to achieve translational invariance. The kernel is used to identify local features but should be invariant of the position of this feature in the image. After convolution is applied, a dimensionality reduction technique like a max-pooling layer is applied, often followed by another convolution layer. The stacking of these two layer-types allows the kernel to learn a hierarchical feature structure and to combine small, local features into more general features and concepts.

Recurrent neural networks rely on a similar kind of translational invariance. The same parameters are used at every time step to search for local features. In addition, the outputs of the previous time step are supplied, as the assessment of the current features might depend on features found at earlier time steps. The initial, recurrent neuron concatenated the previous output (the hidden state) to the input at the current time step and applied the activation function. The repeated usage of the activation function on the hidden state leads to a problem, known as the vanishing gradient problem. This makes it hard for simple recurrent neurons to learn long-term dependencies, which lead to the development of the long-short-term memory cell (LSTM). This cell divides the hidden state into a hidden state and a cell state. The hidden state is used with the input to decide at each time step, which features to use, while the cell state transports the long-term features trough the time step without any activation function being applied. Internally, the LSTM cell uses three activation gates: one forget gate to choose which features of the previous step to use, one update gate to decide how the cell state should be updated and one output gate to decide which features of the state should be used as output. This differentiation makes the LSTM-Cell computational expensive, leading to the development of the Gated Recurrent Unit (GRU). An adaptation using only two internal gates to be a faster version still leveraging the idea of splitting the hidden state and the cell state.

\subsection{Process prediction}
\begin{figure}[ht]
\centering
\includegraphics[width=\textwidth]{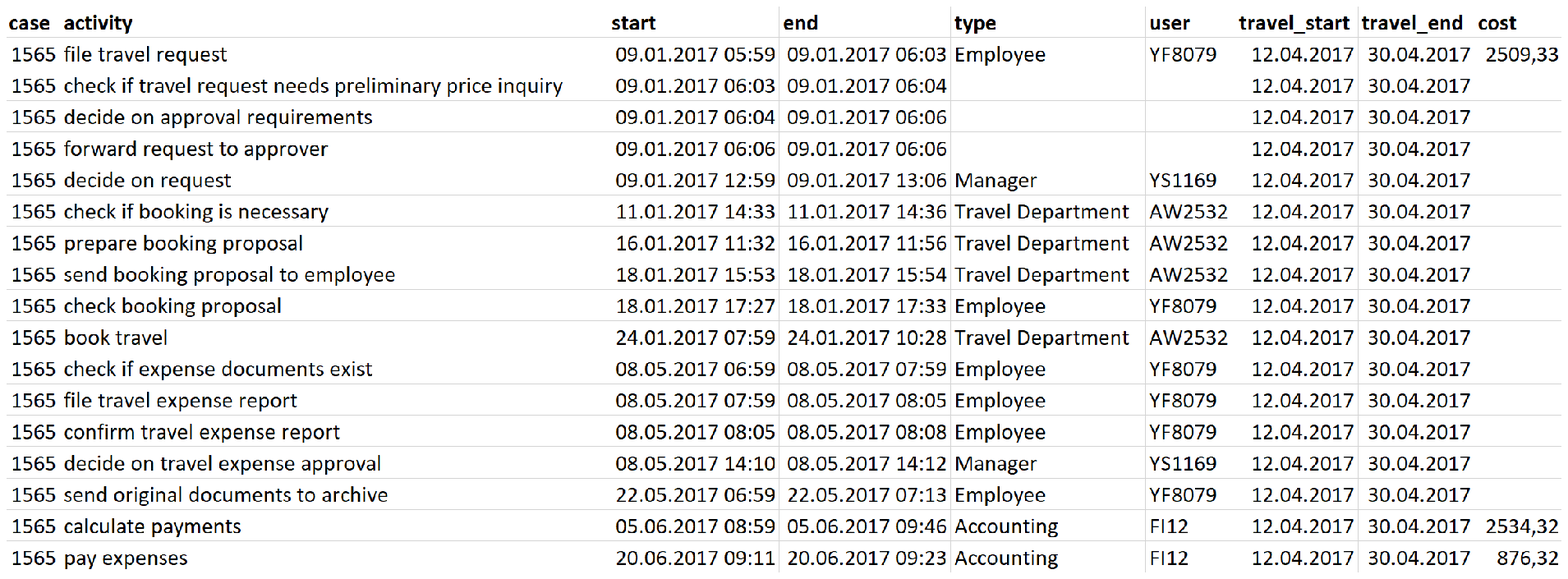}
\caption{Illustration of an event log in tabular presentation}
\label{fig:event_log}
\end{figure}

Process mining analyses event logs created by process-aware information systems (PAIS) to gain insights into the inner logical structure of these business processes. Logs of actual executions are used to get an as-is picture of the process. One execution of a process is called an instance or a trace, which itself consists of a sequence of events. At least three distinct features describe each event. A trace-ID to identify individual executions, an activity-name to identify the different activities that were performed, and a timestamp to express the chronological order and execution time. As additional information, a resource may be attached to an event, to identify the person, group or machine executing the activity. Further attributes describing the activity can be attached. These attributes are also referred to as payload data. In addition, the corresponding process model can contain iteration loops, exclusive branches or concurrent activities. Different versions of the process can be very similar or different each time. Some information refers to a process execution (instance), others to a concrete event. This results in highly divergent event logs. Figure \ref{fig:event_log} shows a cutout from an event log in tabular presentation.

Process prediction, often referred to as predictive process monitoring, focuses on predicting potential outcomes of uncompleted traces. Contrary to standard process mining techniques like process discovery, process prediction is performed during execution and is, therefore, an online algorithm. Process discovery, on the other hand, is often a static task, applied to a set of finished instances. The results are process models of past executions, which serve as a discussion foundation for business process managers. Whereas process prediction is meant to identify unwanted outcomes at an early stage of the process execution, leaving managers with the ability to influence the output through process adaptation.

Consequently, process prediction is performed during the instance execution, i.e. only realized events are known at prediction time. To differentiate between realized events and upcoming unknown events, we split traces into prefixes and suffixes. Prefixes represent events that have already been realized while the suffix depicts the upcoming unknown events. For real-world applications, prediction algorithms should be evaluated on a set of performance measurements, since the business problem at hand defines the required characteristics. In some cases, the mean accuracy is less important than the earliness of the prediction. In other scenarios, earliness is not an important factor. Instead, the predictions have to be highly reliable. Other manifestations of process prediction are the prediction of the next activity, the estimation of the completion time or early detection of abnormal process behavior indicating rule violations or compliance breaches.

\subsection{Deep Learning based Process prediction}
 Given a sequence of events of a running case, the
 objective of process prediction is to forecast how certain aspects of the case unfold in the future. Originally, the focus was to utilize deep learning for next-step prediction and outcome prediction, i.e. to predict the next activity or the last activity of a case based on a sequence of seen activities. This was motivated by the recent success of deep learning for language modeling in the field of natural language processing and its high similarity with process prediction from a conceptional point of view \citep{Evermann.2016}. In fact, the next-step prediction task can be represented with the language modeling formula $P(W_t | W_{t-k}, \dots, W_{t-1})$ by replacing a word $W$ with an activity $A$. For this reason, it is not surprising that some of the concepts that had already been successfully used in the natural language process domain were also applied to process prediction. Examples are one-hot encoding, n-grams, word embedding, recurrent neural networks, etc. Over the past years, other prediction targets such as time-to-outcome, time-to-next-event or the classification of cases into specific groups have been elaborated. 
 
 \begin{figure}[ht]
\centering
\includegraphics[width=0.8\textwidth]{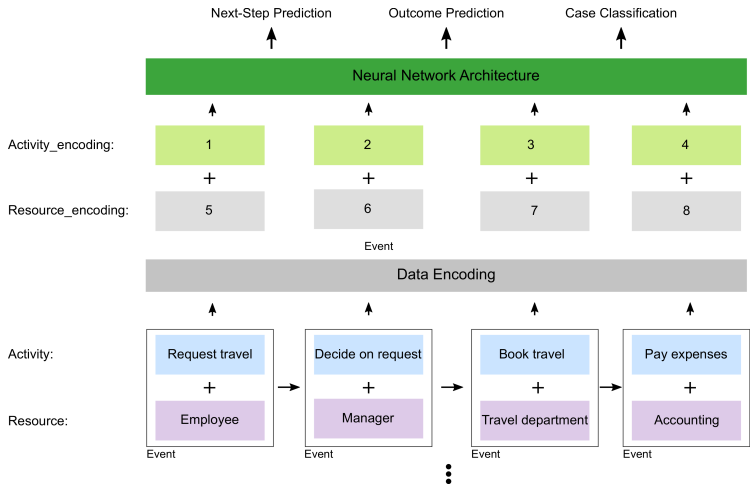}
\caption{Deep Learning based Process Prediction}
\label{fig:dl_based_process_prediction}
\end{figure}

 Figure \ref{fig:dl_based_process_prediction} lays down the basic steps that have to be performed in order to apply process prediction with deep learning.
 In order to pass the activity sequence to the neural network, it has to be converted into numerical representation, first.  If multiple attributes shall influence the prediction, they need to be combined in a suitable encoding. Based on the type of the attributes, i.e. categorical, numerical or time attributes, different encoding schemes are beneficial. Regarding the choice of the encoding and architecture of the neural network, a wide range of options is available, each with its own advantages and disadvantages.
Another important aspect concerns the question how to train the neural network, efficiently. Usually the event log is split into process instances that are used for training the model and process instances that are selected for validation. In some cases, an extra test set is extracted.  In practice, there are often not enough cases to train a deep neural network, properly. Furthermore, the traces consist of a different number of events, which is not supported for training common deep learning architectures. To overcome these issues, many existing approaches extract multiple training samples from one process instance. In addition, the training samples are progressed into fixed length windows of sequences.

\section{Methodology}
To obtain a summarising overview of the research field of process prediction, we conduct a systematic literature review (SLR). To ensure the quality of the literature review, we follow the proposed methodology of \cite{Kitchenham.2009} with a few variations. We start by defining the research questions. Next, we infer the search string needed to query the academic databases. We then apply a set of inclusion and exclusion criteria to filter out studies, not relevant to our research question. Afterwards, we then utilize a forward search to identify relevant papers, not matching the search string to broaden our literature base.
\subsection{Research question formulation}
The goal of this SLR is to synthesise the advantages and disadvantages of procedural decisions when using deep learning for process prediction. These decisions can be broken down into several subcategories.

\begin{itemize}
 	\item[(RQ1)] Which kind of neural network is used for prediction?
	\item[(RQ2)] Which pre-processing steps are carried out? 
	\item[(RQ3)] How is the data being encoded?
	\item[(RQ4)] What is the prediction target?
	\item[(RQ5)] Are there dominant approaches or does every approach come with its own advantages? 
	\item[(RQ6)] Are the proposed approaches combinable?
\end{itemize}

While research questions one to four mainly examine the state-of-the-art, questions five to six deal with the orientation of future research.
Research questions one to four examine the deep learning approach of each paper individually. With RQ1, we analyse the basic structure of the underlying neural network: Does the approach use a feed-forward structure or a recurrent neural network to capture the time dependency. Moreover, we analyse the size of the applied neural networks and their output format. In the next research question, we compare the pre-processing steps performed by the authors. These include activities such as filtering, enhancement or clustering of the log data. Whereas RQ3 focuses on the transformation of the input data to fit the requirements of a neural network. With RQ4, we analyse the different characteristics used as prediction target. These may be a performance measure like execution costs, a class like "success" or "failure" or even a sequence of classes like the whole process-suffix of this trace.    

After the structural conditions of the individual approaches have been examined, they are compared with each other for RQ5, to identify reoccurring and dominant concepts. The final research question will aim to combine the proposed approaches. Grouping existing techniques into fundamentally opposed designs and addable enhancements of those designs makes it easier to find starting points for future research. 
\subsection{Search strings}
The existing literature on process prediction using deep learning was gathered using the common databases for computer science (Web~of~Science, Science~Direct, IEEE~Explore, ACM~Digital~Library \& SpringerLink). 
%Although others used Google Scholar to search all the databases at once; we deliberately chose to search each database individually, which shall be explained later. 

We developed a three-part search string to frame the relevant literature. \textit{"business process *"} was used to get only domain-specific texts. Next, \textit{"predict*"} was used to get the literature concerned with prediction tasks and then \textit{"deep learning"} or \textit{"neural networks*"} were added to specify the used technology. The terms were searched in the title, the keywords and the abstract. The main challenge was the definition of key terms that match a wide area of prediction tasks while focusing on a special technology type. Typically, the authors only handle one prediction task (e.g. suffix prediction or remaining time forecasting). Also, they might use deep learning without explicitly mentioning the words in the title or keywords. On the other hand, searching for \textit{"business process*"} and \textit{"prediction"} in full texts, leads to many false positives in a database query. %\footnote{\textit{"prediction"} AND \textit{"business process"} AND \textit{"deep learning"} results in 3.010 entries (full-text search) or 4 entries (in-title search) in Google Scholar.}. As Google scholar only offers to search the full text or the title, the individual databases were used to specify the search area further.
The adapted search strings are shown in table \ref{tab:search_strings}.

% Eventuell hozizonal in den Anhang oder weg. 
\begin{table}[ht]
    \caption{Search strings and resulting entries}
    \begin{tabular}{p{7em} p{25em} l}
        
        \toprule
        \textbf{Database} & \textbf{Search string} & \textbf{Results} \\
        \midrule
        Web of Science & TOPIC: ("Business Process*") AND TOPIC: (predicti*) \newline AND TOPIC: ("*Neural networks*") & 20 \\
        Web of Science & TOPIC: ("Business Process*") AND TOPIC: (predicti*) \newline AND TOPIC: ("Deep Learning")  & 11 \\
        Science Direct & Title, abstract, keywords: "Business Process"  \newline AND "Predictive" AND "Deep Learning"  & 2 \\
        Science Direct & Title, abstract, keywords: "Business Process"  \newline AND "Prediction" AND "Deep Learning"  & 2 \\
        Science Direct & Title, abstract, keywords: "Business Process"  \newline AND "Predictive" AND "Neural Networks" & 7 \\
        Science Direct & Title, abstract, keywords: "Business Process"  \newline AND "Prediction" AND "Neural Networks" & 5 \\
        IEEE Explore & ((("All Metadata":"Business process*") AND "All Meta- \newline data":Predicti*) AND "All Metadata":"Deep Learning")  & 9 \\
        IEEE Explore & ((("All Metadata":"Business process*") AND "All Meta-  \newline data":Predicti*) AND "All Metadata":"*Neural Networks*") & 27 \\
        ACM Digital Library & Keyword:("Business Process") AND Keyword:(Predicti*) \newline AND Abstract:("Neural Networks") & 0 \\
        ACM Digital Library & Keyword:("Business Process") AND Keyword:(Predicti*) \newline AND Abstract:("Deep Learning") & 0 \\
        SpringerLink & query="neural+networks"+AND+(predict*) \newline \&dc.title="Business+process*" & 36 \\
        SpringerLink & query="deep+learning"+AND+(predict*) \newline \&dc.title="Business+process*" & 14 \\
        \bottomrule
%        & \textbf{Without duplicates} & \textbf{77} \\
%        \bottomrule
    \end{tabular}
    
    \label{tab:search_strings}
\end{table}

Due to the overlap of some databases and the repeated search with similar search strings, 56 duplicates had to be excluded. The next section presents the inclusion and exclusion criteria to further distil the literature.
\subsection{Inclusion and exclusion criteria}
To filter out the relevant studies, we defined the following inclusion and exclusion criteria. A study has to comply with all inclusion criteria and none of the exclusion criteria to be considered in this systematic literature review. 

\textbf{Inclusion Criteria:}
\begin{itemize}
    \item[(IN1)] The study is related to business process prediction and reports an implemented and evaluated approach. 
    \item[(IN2)] The study uses a deep learning method, i.e. some form of neural network for prediction or it combines traditional machine learning algorithms with a deep learning architecture.  
\end{itemize}

\textbf{Exclusion Criteria:}
\begin{itemize}
    \item[(EX1)] The study is not published in English.
    \item[(EX2)] The study is not freely accessible through the standard university libraries proxy services or it is electronically unavailable.
    \item[(EX3)] The study or the presented approach is not related to the field of computer science.
\end{itemize}

After a first screening of the title and abstract, 38 papers where rejected, mostly because they did not relate to business process prediction or implement a deep learning solution. The remaining 39 studies were analysed in-depth, and an additional 19 papers were rejected since they did not implement a new kind of deep learning algorithm, but instead tested existing approaches in a case study. In conclusion, the systematic database search resulted in 21 papers to be reviewed.

\begin{figure}[ht]
\centering
\includegraphics[width=0.5\textwidth]{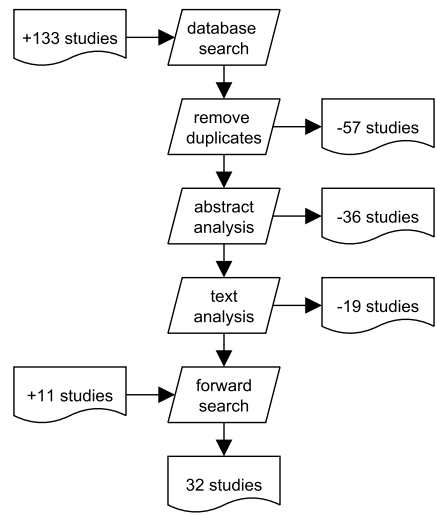}
\caption{Overview of literature identification process}
\label{fig:research_process}
\end{figure}

To further broaden the literature base, we added a one-step forward search on the studies found by the database search. This step is due to the large number of synonyms used by authors. For example, the neural network architecture Long-Short-Term Memory is often abbreviated as LSTM without using the keywords \textit{"deep learning"} or \textit{"neural network"}. Therefore these studies might not have been located by the search strings. The forward search yielded another 11 studies, bringing the total number of studies available for an investigation to 32. An overview of the whole search process is provided in figure \ref{fig:research_process}. The individual references and the exclusion reasons are presented in Online Resource 1.

While having excluded multiple instances of identical publications, we still considered all existing versions of a study, if they contained improvements or new insights. In the analysis, we only refer to the peer-reviewed publication that introduced the approach first. 

\section{Results} \label{sec:result}
This section summarises the results of the in-depth analysis of the relevant studies. In the first step, each approach is dismantled into its fundamental principles of construction. Next, we focus on individual aspects of deep learning process prediction and compare the approaches on a more granular level. After providing a profound examination of the used techniques, we discuss the advantages and drawbacks experienced by the authors in section \ref{sec:discussion} to answer research questions five and six.    

\subsection{Overview}
We provide a high-level overview of the underlying concepts in table \ref{tab:paper_overview}. The general structure follows the research questions one to four, although the preprocessing steps were left out at this point, and we firstly focus on the data used as input. 

The underlying architectures can generally be divided into three different approaches: feed-forward networks (FFNN), convolutional neural networks (CNN) and recurrent neural networks (RNN). A further investigation of the specific recurrent neural network type will be given in section \ref{sec:architecture}. Concerning the prediction target, the studies operate as expected and usually treat only one target at a time. Although many different objectives are pursued, they can be grouped into three distinct types. The target is either to predict the next event, to estimate some performance measure or to classify the entire trace. These three types differ mainly in their underlying mathematical problem formulation, i.e. many different business problems, like anomaly detection, service level agreement violations or the final event of a process, can be subsumed to the same prediction type.

On the input side of the neural network, the approaches show a larger diversion. The inputs can be composed of the four different entities of a business process or handcrafted features thereof. After a subset of these inputs is selected, the variables are encoded, to fit the requirements of a neural network. To transform categorical values and words into appropriate inputs, one-hot-encoding and embedding layers are the dominant strategies. The timestamp is a complicated continuous feature due to its wide scale (from seconds to years). The encoding, therefore, needs special attention and is analysed in more detail in section \ref{sec:encoding}. After this high-level overview, the following sections dive into the procedural decisions made by the authors.

\begin{table}[ht]
    \caption{Overview of the basic concepts for Deep Learning Process Prediction}
    \begin{tabular}{l|*{3}{l@{\hspace{0.8em}}}|*{3}{l@{\hspace{0.9em}}}|*{5}{l@{\hspace{0.8em}}}|*{3}{l@{\hspace{1em}}} }
        \toprule
        \textbf{Study}  
            & \multicolumn{3}{l}{\textbf{Type}}     
            & \multicolumn{3}{l}{\textbf{Target}}        
            & \multicolumn{5}{l}{\textbf{Input}}    
            & \multicolumn{3}{l}{\textbf{Encoding}} 
        \\
        %study title
            & \rot{FFNN}                 & \rot{CNN}             & \rot{RNN}
            & \rot{Next Event}          & \rot{Trace class}     & \rot{Time}
            & \rot{Activity}            & \rot{Ressource}       & \rot{Attribute}   & \rot{timestamp}   & \rot{custom}
            & \rot{One-Hot}    & \rot{Embedding}       & \rot{custom}\\
        \midrule
        \citeauthor{AlJebrni.2018}          &   &\ok&   &\ok&   &   &\ok&   &   &   &   &   &\ok&   \\
        \citeauthor{Bandis.2018}            &\ok&   &   &   &   &\ok&   &   &   &   &\ok&   &   &\ok\\
        \citeauthor{Camargo.2019}           &   &   &\ok&\ok&   &\ok&\ok&\ok&   &\ok&   &   &\ok&   \\
        \citeauthor{DiFrancescomarino.2017} &   &   &\ok&\ok&   &   &\ok&   &   &\ok&   &\ok&   &   \\
        \citeauthor{DiMauro.2019}           &   &\ok&\ok&\ok&   &   &\ok&   &   &   &   &   &\ok&   \\
        \citeauthor{Ezpeleta.2018}          &\ok&   &   &   &\ok&   &   &   &   &   &\ok&   &   &   \\
        \citeauthor{Evermann.2017}          &   &   &\ok&\ok&   &   &\ok&\ok&   &   &   &   &\ok&   \\
        \citeauthor{Hinkka.2019}            &   &   &\ok&   &   &   &\ok&   &   &   &   &\ok&   &   \\
        \citeauthor{Khan.2018}              &   &   &\ok&\ok&\ok&\ok&   &   &   &   &   &\ok&\ok&\ok\\
        \citeauthor{Kratsch.2020}           &\ok&   &\ok&   &\ok&   &\ok&   &\ok&   &   &\ok&   &   \\
        \citeauthor{Lin.2019}               &   &   &\ok&\ok&   &   &\ok&   &\ok&   &   &   &\ok&   \\
        \citeauthor{Mehdiyev.2017}          &\ok&   &   &\ok&   &   &\ok&   &   &   &   &   &   &\ok\\
        \cite{Metzger.2018}                 &   &   &\ok&\ok&\ok&   &\ok&   &\ok&   &\ok&   &   &   \\
        \cite{Metzger.2019}                 &   &   &\ok&   &\ok&   &\ok&\ok&\ok&\ok&   &\ok&   &   \\
        \citeauthor{Navarin.2017}           &   &   &\ok&   &   &\ok&\ok&\ok&   &\ok&   &\ok&   &   \\
        \cite{Nolle.2016}                   &\ok&   &   &   &\ok&   &\ok&   &   &   &   &\ok&   &   \\
        \cite{Nolle.2018}                   &   &   &\ok&\ok&   &   &\ok&   &\ok&   &   &   &\ok&   \\
        \citeauthor{Park.2020}              &   &\ok&\ok&   &   &\ok&   &   &   &   &\ok&   &   &   \\
        \citeauthor{Pasquadibisceglie.2019} &   &\ok&   &\ok&   &   &\ok&   &   &\ok&\ok&   &   &\ok\\
        \citeauthor{Schoenig.2018}          &   &   &\ok&\ok&   &   &\ok&\ok&\ok&   &   &\ok&   &   \\
        \citeauthor{Tax.2017}               &   &   &\ok&\ok&   &\ok&\ok&   &   &\ok&   &\ok&   &   \\
        \citeauthor{Taymouri.2020}          &   &   &\ok&\ok&   &\ok&\ok&   &   &   &   &\ok&   &   \\  
        \citeauthor{Theis.2019}             &\ok&   &   &\ok&   &   &\ok&   &\ok&   &\ok&   &   &   \\
        \citeauthor{Wahid.2019}             &\ok&   &   &   &   &\ok&\ok&   &\ok&\ok&   &\ok&\ok&   \\
        \citeauthor{Wang.2019}              &   &   &\ok&   &\ok&   &\ok&   &\ok&   &   &\ok&   &   \\
        
        \bottomrule
    \end{tabular}
    \label{tab:paper_overview}
\end{table}

\subsection{Network architecture} \label{sec:architecture}
Deep neural networks are experiencing a steep increase in their use for prediction tasks. Researchers in many domains exploit their ability to map many complicated relations. In result, different variations of the original perceptron have been proposed to master various prediction tasks \citep{LeCun.2015}.

Due to the temporal arrangement of the activities in a trace, the authors seem to prefer recurrent architectures as these have been specifically designed to cope with temporally depended data. Whereas \cite{Nolle.2018} and \cite{Hinkka.2019} used Gated Recurrent Units, the majority decides on Long-Short-Term Memory Cells. \cite{AlJebrni.2018,DiMauro.2019,Pasquadibisceglie.2019,Park.2020} on the other hand chose convolutional neural networks due to their simplicity and computational efficiency. 8 studies decided to use feedforward networks. Like CNNs, these FFNNs suffer from a fixed input size forcing the authors to rescale prefixes of different lengths to a fixed size input matrix. \cite{Nolle.2018} and \cite{Pasquadibisceglie.2019} solved this restriction by padding all shorter prefixes to the maximum possible length. \cite{Ezpeleta.2018} trained a separate model for multiple prefix lengths and \cite{Mehdiyev.2020} used n-grams as inputs to model the temporal dependencies while preserving a fixed input size. \cite{Theis.2019} mined a petri net for the event logs and use this information to model the control-flow within an FFNN input.

\cite{Park.2020} even combined the CNN approach with an LSTM architecture using a long-term recurrent convolution network (LRCN), outperforming their LSTM and CNN alternatives. \cite{Khan.2018} used a unique architecture of a differentiable neural computer. This DNC utilizes an external memory to store information. This store benefits from being order-free as opposed to the internal state of an LSTM-cell which is order-dependent due to its sequential design. \cite{Khan.2018} stated, that this feature simplifies the exploitation of long-term dependencies in a process model. Otherwise, all long-term features of a process model have to be represented by the internal LSTM state. \cite{Lin.2019} focussed on modelling the relationship of activities to their attributes. Both are treated separately as inputs and a form of attention layer is used to combine both hidden representations before the predictions are encoded, leading to mayor performance improvements to prior works.

\cite{Taymouri.2020} proposed a generative adversarial architecture (GAN), where one neural network (generator) produces traces from random noise and another network (discriminator) has to differentiate between real traces and the ones created by the generator. This technique originates from the domain of computer vision, where GANs are used to create new images, that resemble real images. Surprisingly, the authors fed real prefixes to the generator and used it afterwards to predict real next events. It is unclear, why \citeauthor{Taymouri.2020} relied on generative adversarial architecture. Generators were designed to create images that look like real images but are not. In their setting, the generator performs a classical supervised classification task.

In terms of generalization capability, it is important to keep the model as simple as possible. The complexity of a deep learning model can be measured by counting the trainable parameters. The initial proposal of \cite{Evermann.2017} had approx. 504.000 trainable parameters on the BPIC12 dataset. The enhanced LSTM model of \cite{Tax.2017} used fewer neurons and no embedding layer, which led them to approx. 208.000 parameters. The FFNN of \cite{Mehdiyev.2017} only trained 55.600 parameters on the same dataset. CNN approaches share the same parameters over a wide range of input. So even though \cite{DiMauro.2019} used 9 hidden layers (3x3) with 32 filters each, the resulting model only has 48.400 parameters.

\subsection{Data preprocessing \& feature engineering} \label{sec:encoding}

To represent business processes in a unified form, the XES standard for process aware information systems and data mining software is used \citep{XES.2016d}. Since this representation does not comply with the input requirements of neural networks, authors are required to pre-process the log files and convert them to a suitable format. Most authors do not specify if they cleaned the data set. This might be due to the fact that deep learning methods are relatively robust to noise compared to existing approaches \citep{Khan.2018}. \cite{Nolle.2016} stated that only completed traces where considered. \cite{Lin.2019} discarded all traces with less than five activities, because these cases were irrelevant for a prediction. \cite{Kratsch.2020} excluded all attributes with an occurrence of below 99\% to avoid additional bias due to excessive fill values.

The implementation of many architectures requires the input features to be of a fixed size. Hence, the authors cut the traces into fixed-sized prefixes \citep{Evermann.2017,Tax.2017,Nolle.2018,Schoenig.2018,AlJebrni.2018,DiMauro.2019,Camargo.2019}. For the prediction of events having a shorter prefix, the feature matrix is padded with zeros. \cite{Taymouri.2020} on the other hand dropped samples, when the prefix was to short. This leads to incomparable results. On the Helpdesk data, for example, they reported their accuracy for prefixes of size 4 and 6. With a mean trace length of 3.6, more than half of the data was dropped in their approach.  

A business process trace consists at least of an activity and a timestamp, but not all the information available in the log was used as input. Most studies used the activity to map the control-flow, while only \citep{Bandis.2018,Camargo.2019,Wahid.2019,Tax.2017,Metzger.2019,Pasquadibisceglie.2019,DiFrancescomarino.2017,Metzger.2017,Metzger.2017b} used the timestamp information as additional input. Resource information was utilized in 5 publications. \cite{Evermann.2017} evaluated the information gain of adding resource information to the network inputs. They concluded that the performance delta depends strongly on the characteristics of the business process. 

The majority of publications used the raw information of the log. \cite{Wahid.2019} did simple enhancements to the raw data, by adding the elapsed time since the start to the features. \cite{Metzger.2018} included information about process instances being executed in parallel into the trace data. This information is available at prediction time and might be important for time predictions if many traces are using one resource simultaneously. \cite{Bandis.2018} drew upon domain knowledge to map the timestamp to binary variables signalling peak hours. \cite{Theis.2019} added a complete preprocessing step and firstly mined a petri net from the event log. The attributes of this petri net were used afterwards to make deep learning predictions. \cite{Park.2020} followed a similar approach and mined an annotated transition system prior to their actual deep learning prediction. \cite{Ezpeleta.2018} applied linear-temporal-logic rules to the trace and used the resulting signatures as input for their FFNN to perform classification.

For the usage in a neural network, most inputs need further processing. The inner logic of neurons is designed for input vectors with values between -1 and 1 and a mean of 0. The process of adjusting the input data to this range is known as data encoding. Scalar values are usually encoded with a min-max-normalization or their standard-score. Categorical data can be encoded in multiple ways: one-hot-encoding or embedding are two possibilities. For both, the category is first converted into an integer between 0 and number of categories - 1. A one-hot-encoding results in a vector of size \textit{number of categories}. All values are zero except the value at the index of the category, which is one. An embedding, on the other hand, maps the index to a vector of arbitrary length with real-valued values. Embeddings are used as a dimension reduction technique when the number of categories is large. Additionally, they are able to provide a kind of proximity measure for the classes due to the real-valued data. One-hot-encoded classes are treated as equidistant. 

Both encoding techniques are used throughout the publications. Eleven studies fell back on one-hot-encoding, and another 11 studies used embedding. \cite{Wahid.2019} directly compared one-hot-encoding to embedding in their implementation and found that embedding leads to increased performance. \cite{Tax.2017} observed better performance with their one-hot-encoding solution compared to the embedding approach of \cite{Evermann.2017}. However, their approaches differed in more design choices than the encoding, making the results hard to pinpoint on the encoding. Additionally, the vocabulary size (number of categories) on the underlying dataset was 7, resulting in a one-hot-encoded vector of size 7, while \cite{Evermann.2017} used an embedding size of 500. So for this dataset, the embedding did not perform a dimensionality reduction, but rather a heavy dimensionality expansion.

The conversion of the time stamp into a suitable format requires more case-specific decisions. Treating the time as single scalar, the standard score could be used. However, this representation would make it hard to correlate seasonal dependencies. Business processes can have seasonal changes at many levels (per-month, per-week, per-day) and the time interval between events is heavily dependent on the business process. Therefore, a universal time-encoding is not possible. \cite{DiFrancescomarino.2017} used a 3-fold time encoding with three scalars: relative time increase since the last event, time since midnight and time since begin of week. \cite{Pasquadibisceglie.2019} expressed the execution time in days as a single scalar. The authors do not mention if a normalization was applied before training, as the scalar exceeds the usual input size of neurons by at least one magnitude. \cite{Navarin.2017} used the time from trace start, time from the last event, the time and weekday in which the event started, but also do not mention any normalization. \cite{Camargo.2019} used the relative time since the last activity and applied min-max-normalization. However, because some executions took a very long time, the distribution of the variable is highly skewed. So they try log-normalization as well and observe better performance with log-normalization when the time interval is highly variable.

\subsection{Prediction target}

To categorise the analysed prediction targets, we differentiated the tasks along two dimensions: the prediction level and the prediction type. The former category splits into process-level predictions and time-step level predictions. A prediction on a process level can be carried out at any time step. However, it is always expected to yield the same result, i.e. a characteristic of the entire trace. In contrast, a timestep level prediction should yield results for the upcoming timestep. It is therefore expected to get different results at every prediction time step. This differentiation is made because the performance evaluations of these two methods diverge. As \cite{Teinemaa.2018} pointed out, the value of early process level predictions is higher than the latter ones, and their usability deteriorates when classification results bounce back and forth. As a result, predictions on a process level need to be evaluated on their earliness and temporal stability.

The grouping of existing approaches by their underlying prediction type focuses on the subsequent evaluation metrics. We distinguish between regression and classification tasks. In both cases, we use a performance metric to calculate the distance between the prediction of the model and the actual value. For regression tasks, mean squared error (MSE) or mean absolute error (MAE) are standard metrics, whereas accuracy (Acc.), precision (Prec.) or recall are used in classification problems.

\begin{table}[ht]
    \caption{Classification of prediction targets}
    \begin{tabular}{p{7em}p{14em}p{14em}}
        \toprule
        \textbf{}       
            & \textbf{Process level}                                                    
            & \textbf{time-step level}\\
        \midrule
        Regression      
            & Process duration [2]                                                      
            & Remaining time [1]\newline Time to next event [3]\\
        Classification  
            & Final event [2]\newline Process classification [5]\newline Anomalistic process [1]  
            & Next event [16]\newline Next resource [3]\newline Anomalistic activity [1]\\
        \bottomrule
    \end{tabular}
    \label{tab:prediction_classes}
\end{table}

Table \ref{tab:prediction_classes} shows the focus on next-event prediction and process classification. For regression targets, only time is considered. Even though other scalar values might be of interest for business applications, we found no publication that focuses on execution cost or attribute value estimation. Also, few studies are concerned with a prediction on a business process model level. \cite{Park.2020} state that predictions on an instance level are useful for short term process instance adaptations, but business process managers are primarily concerned with business process model optimizations. Hence the authors predict the mean activity execution time for all running instances at a given time step. 

Additionally, time performance predictions using only one trace as input implicitly presume that the depended variable is in the trace itself. However, the workload of each resource can strongly influence the execution time, i.e. the number of concurrent instances also using this resource. As \cite{Metzger.2018} point out, this information is available at prediction time and does not require a preceding process discovery step.

It can also be seen from table \ref{tab:paper_overview}, that the studies mainly focus on single prediction targets. Only \cite{Metzger.2018,Khan.2018,Tax.2017} performed multiple predictions. \cite{Metzger.2018} compared an iterative next-event prediction to a direct final event prediction approach, whereas \cite{Khan.2018,Tax.2018} leveraged multitask learning by predicting a time measure and the next event with the same neural network. Neural networks can be trained on multiple tasks simultaneously. This is achieved by using multiple output heads, each responsible for a different prediction. The usage of multiple, inter-related tasks can help the network to leverage the additional information in each of the tasks to improve the generalization capability on all of the tasks. \citep{Zhang.2017}

To compare prediction results on each task, not only the architectural choices are relevant, but also the datasets used for evaluation—noise and available information in the event log influence the achievable results. Therefore, we present a summary: Overall, the evaluation of the results is done on 28 different datasets. Most approaches are only evaluated on a small subset of the publicly available process logs. This heterogeneity in test data makes the studies hard to compare. The most prominent process logs are the business process intelligence challenge from 2012 with 18 usages, followed by the Helpdesk dataset with 13 usages. \cite{Nolle.2018,Wang.2019,Lin.2019,DiFrancescomarino.2017,Hinkka.2019,Khan.2018} evaluated their implementation on five datasets or more to study the robustness of their approach while 14 publications only use one or two process logs to test.

In the subsequent paragraphs, we will present the individual performance measures reported by the authors. The tables should be used to get a holistic impression of the capacity of current deep learning prediction algorithms. For roughly comparable results, we extracted the single scalar measurements by following a structured process. Firstly, if the paper reported the results on one of the most commonly used datasets, we chose these datasets, even if the approach yielded higher scores on different datasets. If none of the two datasets was used, we firstly report the results on other BPI Challenge datasets, and then we fell back on uncommon or private data. On the chosen dataset, we selected the best variant, even if this variant did not dominate on all datasets. When per-class metrics were reported, as in \cite{Ezpeleta.2018}, we display the arithmetic mean.  Most authors executed the performance evaluation on a separate testing dataset to mitigate overfitting. Only \cite{Nolle.2016} seems to have evaluated the approach on the training data. When splitting data into testing and training sets, random sampling or strategic partitioning are possible policies. 

On the other hand, k-fold cross-validation uses random sampling to create k subsets. The training is performed multiple times, and at each iteration, a different fold is used as a testing set. The final score is the arithmetic mean of all k tests. This method is more robust as all the available data points are used once in the test sample to represent the entirety. Furthermore, smaller folds (larger k) lead to larger training sets and potentially better scores, but smaller folds are computationally more expensive, as the training has to be performed k times. Therefore, we included the testing method, as it may influence maximum performance and metric reliability.            

\begin{figure}
    \centering
    \includegraphics[width=10cm]{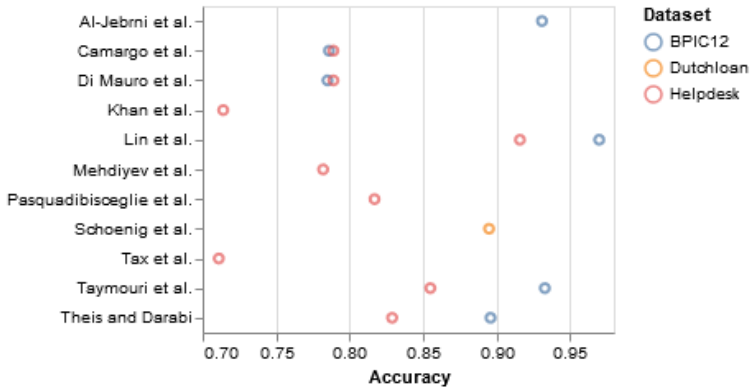}
    \caption{Individual results on next-event prediction}
    \label{fig:results_nextevent}
\end{figure}

Figure \ref{fig:results_nextevent} shows the results for the next-event prediction task. Most authors report their prediction accuracy, i.e. the percentage of correctly classified activities. Unfortunately, this metric is susceptible to class frequency, which will be discussed in more detail in section \ref{sec:discussion}. The Matthews correlation coefficient (MCC) is similar to the accuracy and less susceptible to class frequency. The precision captures type \uppercase\expandafter{\romannumeral1\relax} errors, whereas the recall is focusing on type \uppercase\expandafter{\romannumeral2\relax} errors. Contrary to the accuracy, the F1-Score presents a harmonic mean of precision and recall by balancing the two.
AUC is the area under the receiver operating characteristic curve.The results of \cite{Evermann.2017} are not shown in the figure, since the authors only reported the precision.

For the task of next-event prediction, current top architectures reach an accuracy of 90\% or higher. Interestingly, the most promising studies each follow a different approach. \cite{Lin.2019} used the combination of LSTM layers and a modulator layer to create a form of attention. \cite{AlJebrni.2018} relied on a deep convolutional neural network and \cite{Theis.2019} expanded the data preprocessing step by mining a petri net in advance.

\begin{figure}
    \centering
    \includegraphics[width=10cm]{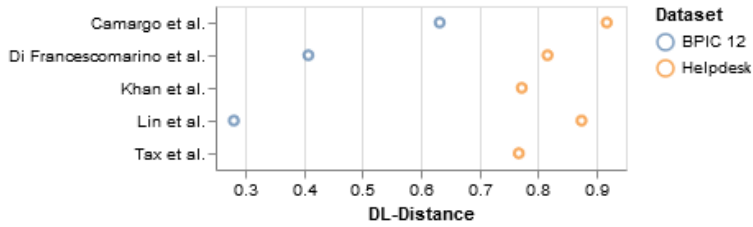}
    \caption{Individual results on suffix prediction}
    \label{fig:results_suffix}
\end{figure}

While 14 studies performed next-event prediction, only 5 evaluated the ability to predict all upcoming events until process completion. The measurement in use throughout the studies is the Damerau-Levenstein (DL) distance or more precisely, a normalized similarity based on the DL distance. The Levenstein distance counts the number of insertions and deletions of activities to convert the predicted suffix into the correct suffix. The Damerau-Levenstein distance additionally allows a swapping operation instead of one insertion and one deletion. The delta between the results on the BPIC 12 and the Helpdesk data underline the importance of the event log in terms of maximal achievable performance (see figure \ref{fig:results_suffix}). The dependent variables for this prediction task might not be in the event log itself.

\begin{comment}
\begin{table}[ht]
    \centering
    \caption{Individual results on suffix prediction}
    \label{tab:results_suffix}
    \begin{tabular}{llll}
        \toprule
        \textbf{Study}                      & \textbf{Dataset}  & \textbf{DL-distance}  \\
        \midrule
        \citeauthor{Camargo.2019}             & BPIC 12 & 0,632 \\
                                              & Helpdesk& 0,917 \\
        \citeauthor{DiFrancescomarino.2017}   & BPIC 12 & 0,408 \\
                                              & Helpdesk& 0,816 \\
        \citeauthor{Khan.2018}                & Helpdesk& 0,772 \\
        \citeauthor{Lin.2019}                 & BPIC 12 & 0,281 \\ 
                                              & Helpdesk& 0,874 \\ 
        \citeauthor{Tax.2017}                 & Helpdesk& 0,767 \\ 
        \bottomrule
    \end{tabular}
    
\end{table}
\end{comment}

The trace classification results are shown in table \ref{tab:results_classification}. Although most authors performed a binary classification, the underlying business problems are different and might be of varying difficulty. \cite{Nolle.2018} identified anomalistic traces, whereas \cite{Metzger.2018} predicted final events and \cite{Ezpeleta.2018} maped new traces to known process clusters. \citeauthor{Metzger.2018} only reported the MCC (0,626) and are therefore not listed in the table.

\begin{figure}
    \centering
    \includegraphics[width=10cm]{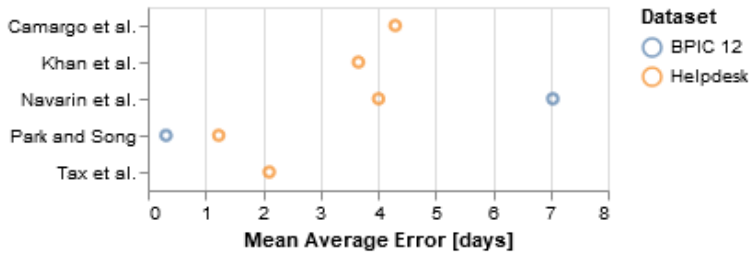}
    \caption{Individual results on performance prediction}
    \label{fig:results_performance}
\end{figure}

When analysing scalar process performance measures, the results get even harder to compare, because the evaluation metrics are not scaled to a fixed interval. In this case, a meaningful comparison can only be made on identical data (see figure \ref{fig:results_performance}). For future evaluations, scaled metrics should be used, e.g. the coefficient of determination. The regression results of \cite{Bandis.2018} \& \cite{Wahid.2019} are not shown, since their evaluation was done on individual datasets with time scales, several magnitudes smaller. 

\begin{table}[ht]
    \centering
    \caption{Individual results on trace classification; *values estimated from figure}
    \label{tab:results_classification}
    \begin{tabular}{llllllll}
        \toprule
        \textbf{Study}          & \textbf{Dataset}  & \textbf{Acc.} & \textbf{Prec.}    & \textbf{Recall}   & \textbf{F-Score}  & \textbf{ROC}  \\
        \midrule
        \citeauthor{Ezpeleta.2018}  & BPIC 11   & -             & 0,972         & 0,97          & 0,97  & -     \\
        \citeauthor{Kratsch.2020}   & BPIC 11   & 0,750         & -             & -             & 0,847 & 0,729 \\
                                    & BPIC 13   & 0,927         & -             & -             & 0,917 & 0,923 \\
        \citeauthor{Metzger.2017}*  &  Cargo    &  $\sim$ 0,93  & $\sim$ 0,99   & $\sim$ 0,69   & -     & -     \\
        \cite{Nolle.2016}           & Synthetic & -             & 1             & 1             & 1     & -     \\ 
        \cite{Nolle.2018}           & BPIC 12   & -             & -             & -             & 0,58  & -     \\
        \citeauthor{Wang.2019}      & BPIC 12   & -             & -             & -             & -     & 0,693 \\
        \bottomrule
    \end{tabular}
\end{table}

\begin{comment}
\begin{table}[ht]
    \centering
    \caption{Individual results on performance prediction}
    \label{tab:results_performance}
    \begin{tabular}{lll}
        \toprule
        \textbf{Study}              & \textbf{Dataset}    & \textbf{Mean average error}  \\
        \midrule
        \citeauthor{Bandis.2018}    & private       & 64,4 [secs]   \\
        \citeauthor{Camargo.2019}   & Helpdesk      & 4,3 [days]    \\
        \citeauthor{Khan.2018}      & Helpdesk      & 3,66 [days]   \\
        \citeauthor{Navarin.2017}   & BPIC 12       & 7,04 [days]   \\
                                    & Helpdesk      & 4,01 [days]   \\
        \citeauthor{Park.2020}      & BPIC 12       & 0,315 [days]  \\
                                    & Helpdesk      & 1,23 [days]   \\
        \citeauthor{Tax.2017}       & Helpdesk      & 2,11 [days]   \\
        \citeauthor{Wahid.2019}     & Port logistics& 190,32 [mins] \\
        \bottomrule
    \end{tabular}
\end{table}
\end{comment}

After the individual results have been presented, the next section will synthesise and discuss the findings of the studies. We focus on insights that should impact future research on all prediction targets. 
\section{Discussion} \label{sec:discussion}
This section is structured as follows. We start by discussing the input data and the corresponding pre-processing steps. Next, the recurrent architectures identified in this review will be analysed in depth. This will unveil a common problem of deep neural networks, known as vanishing gradients. Especially stacked LSTM layers suffer from this problem. As a consequence, we will consider the advantages and disadvantages of convolutional neural networks. Afterwards, we change the perspective and discuss the class frequency problem when doing trace classification or next event prediction. We conclude with a general comparability assessment of the studies.   

\subsection{Consideration of input features}
Throughout the studies, the usage of more input features lead to better performance. There were some exceptions, where adding new information resulted in a deteriorated performance, but in general, the opposite turned out to be true. Surprisingly, only one publication made use of all four entities available in a process log (activity, timestamp, resource and attributes). Further research might evaluate possible ways to address input selection without domain knowledge. In addition, the sparsity of attributes should be investigated, when attributes only relate to a subset of all activities. \cite{Kratsch.2020} mention the risk of additional bias when using excessive fill values, but they do not investigate if this assumption is valid.

Furthermore, the use of deep learning methods enables process prediction algorithms to utilitize new kinds of attribute data. This might be sensory data from a machine used as a resource, the textual description of a problem attached to the attributes in the Helpdesk data or images from cameras monitoring the process execution. The latter two usually need much deeper neural networks to extract useful low dimensional features. However, in the spirit of transfer learning, pre-trained networks can be attached to the process prediction network. 

The review of encoding methods for categorical methods did not yield a definite result. Embedding is usually used for wordlists with a vocabulary size from thousands to millions. For small categorical data with 5 to 20 classes, the benefits of embedding might be negligible. For larger scales, embedding should be able to play out its superiority. Further research would be interesting for categories that can be logically allocated to clusters. Like resources that can be aggregated to resource groups, which therefore have a meaningful distance measure between each other. In such cases, dimensionality reduction should be possible without information loss.

%\textbf{Synthetic logs}
When evaluating their architectures, all studies report different results for each log. This indicates that process specific characteristics influence the performance of the model.
Only a few analyse the possible reasons for this behaviour. It is common in other deep learning disciplines, such as computer vision to use synthetic input data to analyse the behaviour of models. Therefore, synthetic event logs could help to unveil the problematic characteristics of an event log. Other researchers could use this information to focus on specific process characteristics when implementing new architectures. Identified problematic characteristics include loops in the control-flow, repeated consecutive executions of the same activity and activities executable in parallel. 

\subsection{Ellaboration on network architecture}
This review shows the dominant use of a recurrent structure to model time-dependency. Surprisingly, most authors use a fixed prefix size throughout the training. This might be due to computational restrictions, but in theory, recurrent networks such as the LSTM are able to process prefixes variable in length during training. For example the current version of TensorFlow (2.1) \citep{Tensorflow.2015} allows different prefix sizes for supervised training. Although for computational reasons, prefixes of a training batch have to be the same, which can be achieved by online-padding. This way, no prefix has to be cut or aggregated with the threat of information loss and computational effort is reduced, compared to padding all prefixes to the maximum possible size. Another advantage of this dynamic behaviour of RNNs reveals itself in case a new prefix is longer than the maximum prefix in the training set. In this situation, RNNs can still perform predictions, while models with a fixed input size and paddings like CNNs or FFNNs would not be able to predict an outcome in this case.

\cite{Khan.2018} suggest the use of external storage for better long term memory. They tested their architecture on the BPIC 12 and the Helpdesk data with mean traces lengths of 20 and 4,7 respectively. Still, the architecture of \cite{Lin.2019} outperformed their approach with a regular LSTM network. This might imply, that the sequence length seen in BPIC 12 and the Helpdesk data is not a problem for the long-term memory of an LSTM. Results should be compared on much longer sequences such as the Click-Log dataset (BPIC 16) with up to 9.701 events in one trace. On these long sequences, the LSTM architecture of \cite{Lin.2019} achieved its worst results.

%\textbf{vanishing gradient and multitask learning}
The internal computation of LSTM Gates uses tanh and sigmoid activation functions. These activation functions have been proven to cause the vanishing gradient problem when backpropagating through deep neural networks. Especially when stacking multiple layers of LSTMs, training becomes more difficult. Gates of LSTM need to scale their output to the range of -1 to 1, so other activation functions like ReLU are not applicable. To facilitate the learning process throughout deeper recurrent networks, LSTM should be supplemented with simple perceptrons or convolutional layers. This way, higher-level features can be calculated without the problem of vanishing gradients.

%\textbf{Convolutional Networks} 
Convolutional networks have proven to be very powerful in recognising small and large feature patterns in their inputs. While the approach of \cite{Pasquadibisceglie.2019} relies on handcrafted features, \cite{AlJebrni.2018} show that embedding raw data and convolutional networks can be combined. Their 11 layer network demonstrates the capability of deep neural networks to learn elaborate patterns in process logs. Their approach reached 93\% accuracy on the BPIC 12 dataset, which is close to the current maximal performance of the LSTM architecture of \cite{Lin.2019}, reaching 0.97\% accuracy. Additionally, \cite{Park.2020} demonstrate that convolutional and recurrent layers can be combined into one model, outperforming their individual counterparts. Through the use of global pooling layers, CNNs would additionally break their fixed input size restriction, paving the way for new input types.

\subsection{The issue with imbalanced class frequencies}
A common problem in deep learning classification tasks is the relative class frequency. For optimal prediction results, the prediction classes should be equally frequent. When frequencies diverge, deep learning classifiers tend to overfit to the dominant classes and perform worse on the infrequent classes. Because most publications do not evaluate the per activity precision and recall metric, only \cite{Pasquadibisceglie.2019} noticed the strong correlation between class frequency in the training set and class accuracy of the model. \cite{Metzger.2018} used Matthews correlation coefficient instead of accuracy because the accuracy is susceptible to imbalanced data, but they do not fit their training to this insight. \cite{Mehdiyev.2020} compared their aggregated prediction results on balanced and imbalanced training data and report a steep increase in performance, when the training data is balanced. The most commonly used datasets Helpdesk and BPIC 12 are heavily imbalanced. A naive model that predicts only the final event \textit{closed} reaches 37\% accuracy on the Helpdesk data. While important events such as the \textit{raise support level} activity show a frequency of $< 1\%$. For business managers, this infrequent activity is likely to be cost and time-intensive. The prediction accuracy of small frequency activities can therefore be of utmost importance for practitioners. To cope with this prediction problem, over and under sampling are used in deep learning applications to balance datasets during training. The effects of these techniques should be analysed in more detail.

\cite{Nolle.2016} used the property of imbalanced data in their approach: an autoencoder is trained to reconstruct a completed trace. As reconstruction error, the mean squared error between the input and the output is used. After training, they classify a trace as anomalous if the reconstruction error surpasses a certain threshold. The approach works well on their synthetic dataset, where every infrequent deviation of the main process model is considered an anomaly. On real-life datasets, not all infrequent activities or deviations have to be anomalies. In conclusion, to provide reliable predictions in all situations, over and under-sampling of the training data should be considered. Furthermore, if the identification of abnormal or unusual traces is of interest, imbalanced training sets help in the identification of infrequent patterns.

\subsection{Thoughts on quantitative comparability of existing approaches}
The analysed literature on deep learning methods for process prediction lacks quantitative comparability. This is due to several factors. To evaluate the relative performance of publications, one needs a common performance measurement. While most authors rely upon the prediction accuracy, this metric does not capture all relevant aspects. The accuracy is a metric incorporating two kinds of errors: false negatives and false positives. The reduction of one error type always yields an increase of the other, so in classification tasks, an appropriate balance between these two is desired. In business applications, however, one error type can be more expensive than the other one, e.g. when predicting service-level-agreement violations. This is the reason why presented approaches should be analysed on both prediction errors individually and their sensitivity when focusing only on one error type. Additionally, the accuracy score is susceptible to imbalanced datasets, i.e. naive classifiers can reach higher accuracy scores on imbalanced data. This makes the accuracy a problematic metric when it is used on different testing data. 

In addition to balancing the data set, the choice of inputs also plays a significant role in the maximal achievable performance scores. This makes it infeasible to differentiate between the performance effect of the architectural choices and the input choices. Furthermore, if the authors did not perform an exhaustive hyperparameter search, their core contribution could have yielded a higher performance metric. In terms of the testing procedure, we also identified different strategies. Some authors used k-fold validation of varying sizes, others used a fixed percentage as testing data, and some papers even tested on the training data. We propose to use a unified evaluation approach for further publications like k-fold validation, as its reliability has been proven for small and big datasets. A unified benchmark of existing implementations on the wide range of publicly available data would be required to allow for a trustworthy quantitative assessment. Moreover, it would be useful to entail additional metrics, e.g. the per-activity prediction accuracy or the temporal stability of trace-level predictions, to capture all of the relevant aspects of the process prediction.

\section{Future Challenges \& Research Agenda}

In section \ref{sec:discussion}, we discussed the methodological decisions made by authors in current approaches. Based on this evaluation, we presented potential next steps to ameliorate the results of these approaches. In this section, we take a step back and provide a more general view on the field of process prediction. During our evaluation, we identified five major challenges for business process prediction that should be addressed in future research: 
\begin{enumerate}
    \item \textbf{Prediction to action:} the need for process prediction is often justified through the ability to identify malicious or unwanted process executions at an early stage. The economic benefit is achieved through process adaptations at runtime, that might mitigate unwanted process behavior. Hence, the process prediction is only a first step, and the value of the prediction is defined by the performance improvements achievable through its corresponding process adaptation. The nature of the business process provides a wide range of possible adaptations: change the next step to be executed, change the resources allocated to the next step, increase the execution priority and many other. However, the translation of a prediction result into an appropriate adaptation is no trivial task.
    
    In our study, we found that only \cite{Metzger.2019,Metzger.2017} investigated their prediction quality through the results of the consecutive process adaptations. They revealed the important trade-off between earliness and accuracy of prediction, since the implementation of adaptations usually has non-negligible latencies. Future approaches should therefor incorporate the conversion of predictions into action suggestions, as this holistic view might highlight further requirements for a prediction.
    
    \item \textbf{Leveraging domain knowledge:} process mining is applied to gain prescriptive insights on processes through the analysis of event logs. While this data-centric approach yields scalable and transferable methods, it neglects the predictive information in human knowledge. The latter can be used at training time or during predictions to enhance and support data-based algorithms. 
    
    The identification of concurrent events, for example, is extensively studied and hard to detect, even in large datasets. Human experts on the other hand can identify those with little to no effort, since they know the events are generated in distinct systems or they might be handled by different departments. This concurrency information could be leveraged at training time through data modelling. 
    
    As an example, in the permit log of the BPIC 20 dataset, there are activities relating to a travel permit instance (\textit{Permit APPROVED by ADMINISTRATION}) and multiple travel declaration instances (\textit{Declaration SUBMITTED by EMPLOYEE}). By combining the textual description with a brief log analysis, one can conclude that multiple declarations can be submitted independently and that their respective events are concurrent. This information is utilized at training time, by representing the events not as a sequence, by as an acyclic directed graph.
    
    Another exemplary case of useful domain knowledge could be the temporal unavailability of one resource due to technical issues. Such information is rarely present in event logs but is highly relevant at prediction time for short-term process adaptations. \cite{DiFrancescomarino.2017} exploited such rules and thereby improve their prediction results. Although they did not include real domain knowledge, but Linear Temporal Logic rules extracted from the log, the concept is transferable to real time domain knowledge. Hence, future approaches could increase their performance by developing structured approaches to incorporate human domain knowledge into the prediction process.
    
    \item \textbf{Input variables:} closely related to the usage of domain knowledge is the selection of input variables. Through the ongoing digitization of business processes, the amount of electronically available process-related data is growing, whereas current approaches merely focus on the control-flow, i.e. the sequence of activities. Thereby, a significant part of the contextual information on process executions is dropped even though it may play a vital role in the prediction of future outcomes. This context information can be structured in key-value pairs such as the case attributes in the BPIC 20 (\textit{Overspend, TotalDeclared, BudgetNumber, AcitivityNumber, ProjectNumber}) or may be unstructured as in the Helpdesk process, where the textual description of the ticket could be included as input. Recent advances in Natural Language Processing have shown that deep neural networks are able to extract useful information out of unstructured data. This is an advantage of deep learning algorithms that has not been exploited by current approaches.
    
    At the same time, widening the scope of potential input variables will induce the need for systematic selection methods. Especially, when the explanatory power of an attribute will vary from business case to business case. \cite{Evermann.2017} support this hypothesis, as they observed negative and positive performance deltas when they included resource information as input. The influence differs from event log to event log in their experiment. Consequently, a methodical framework to identify explanatory variables for business processes should be developed. 
    
    Aside the selection of appropriate input data, the representation of events should be revised. Handling events as sequential data implies some dependency, i.e. when one event precedes another, it is a prerequisite of this event.  Consequently, \cite{DiFrancescomarino.2017} show that current approaches struggle with concurrent control-flow branches and loops. Process executions should be represented in a data structure that is capable of representing the concurrency and loop information in a more natural way. We propose the use of directed graphs: the internal logic of a process execution is kept and could be leveraged by future prediction algorithms.
    
    % BPIC 20 mit mehreren Case-Attributen: Overspend, TotalDeclared, BudgetNumber, ActivityNumber, ProjectNumber
    % BPIC 20 Concurrency in Variant 555 -> drei Declarations gleichzeitig eingereicht und gehandled.
    % helpdesk: Textual description des Tickets ist wohl ausschlaggeben für Predictions des Tickets.
    
    \item \textbf{Causality and explainability of prediction:} aside the question of \textit{What will happen?}, future approaches should as well answer the question \textit{Why did it happen?}. It has been shown that the "black-box" nature of neural networks hinders the application of prediction models in real-life scenarios because decision-makers do not trust predictions provided as-is with no further explanation.
    When process prediction is the basis for high-value business judgements or is used for automated decision-making, it is crucial to create a profound understanding of the model's behavior and the logic behind its prediction. Providing a relation of causality benefits the decision support system in two ways. On one side, it creates a prediction refutable by humans, as the causal relationship of input data to output can be scrutinized. Which may lead to broader acceptance in practice. On the other side, identifying root causes for unwanted process behavior may support business process model improvements, thereby eliminating the need for repeated process instance adaptations. The research field of explainable AI developed a range of methods to depict and explain the reasoning of neural networks. These approaches should be transferred to the domain of process prediction. One possibility to depict causality is the usage of relevance scores, provided through the neural network weights. These allow for an importance assessment on the individual input variables. 

    % \item \textbf{Prediction explainability:} there is a need to address the "black-box" nature of deep learning methods in order to make the predictions of deep learning models interpretable to domain experts. It has been shown, that the "black-box" nature of neural networks hinders the application of prediction models in real-life scenarios because decision-makers do not trust predictions, that are provided as-is with no explanation.  The research field of explainable AI developed a range of methods to depict and explain the reasoning of neural networks. These approaches should be transferred to the domain of process prediction. The aforementioned depiction of causality is one possibility to extend the explainability of deep neural networks.
    % Kratsch 2020 nutzt einen synthetischen Review Prozess und sagt, die Prediction über die Annahme würde dem Editor helfen. Genauso kann es aber auch dem Author helfen, wenn ihm begründet wird, warum sein Paper abgelehnt wird.
\end{enumerate}

\section{Conclusion}
In recent years, there is an increasing interest in deep learning for process prediction. Although the first studies were only published in 2016, the research community has presented many new approaches.
In this paper, a systematic literature review is carried out to capture the state-of-the-art deep learning methods for process prediction. In total, 32 different approaches are compared against carefully selected criteria to identify strengths and weaknesses and reveal research gaps for future research.

The main focus of this literature review lays on a qualitative comparison of existing implementations. In particular, the literature is classified along the dimensions: neural network type, prediction type, input features and encoding methods.
Although section \ref{sec:result} also reports on the performance scores as presented in the examined papers, due to the sparse intersection of testing data and evaluation methods, it is impossible to carry out a reliable performance ranking. As discussed in \ref{sec:discussion}, a unified benchmark would allow a quantitative comparison of process prediction approaches.

Deep learning enables new levels of decision support in the context of business process management. We presented a list of approaches with increasing performance measures. In our opinion, it is possible to improve them even further by broadening the scope of research to the five major challenges discussed above and thereby creating new tools to support business process managers in their everyday work.
\bibliographystyle{spbasic}
\bibliography{biblio}

\begin{thebibliography}{45}
\providecommand{\natexlab}[1]{#1}
\providecommand{\url}[1]{{#1}}
\providecommand{\urlprefix}{URL }
\expandafter\ifx\csname urlstyle\endcsname\relax
  \providecommand{\doi}[1]{DOI~\discretionary{}{}{}#1}\else
  \providecommand{\doi}{DOI~\discretionary{}{}{}\begingroup
  \urlstyle{rm}\Url}\fi
\providecommand{\eprint}[2][]{\url{#2}}

\bibitem[{Van~der Aalst et~al.(2010)Van~der Aalst, Pesic, and
  Song}]{Aalst.2010}
Van~der Aalst W, Pesic M, Song M (2010) Beyond process mining: from the past to
  present and future. In: International Conference on Advanced Information
  Systems Engineering, Springer, pp 38--52

\bibitem[{Van~der Aalst et~al.(2011)Van~der Aalst, Adriansyah, De~Medeiros,
  Arcieri, Baier, Blickle, Bose, Van Den~Brand, Brandtjen, Buijs
  et~al.}]{Aalst.2011}
Van~der Aalst W, Adriansyah A, De~Medeiros AKA, Arcieri F, Baier T, Blickle T,
  Bose JC, Van Den~Brand P, Brandtjen R, Buijs J, et~al. (2011) Process mining
  manifesto. In: International Conference on Business Process Management,
  Springer, pp 169--194

\bibitem[{Abadi et~al.(2015)Abadi, Agarwal, Barham, Brevdo, Chen, Citro,
  Corrado, Davis, Dean, Devin, Ghemawat, Goodfellow, Harp, Irving, Isard, Jia,
  Jozefowicz, Kaiser, Kudlur, Levenberg, Man\'{e}, Monga, Moore, Murray, Olah,
  Schuster, Shlens, Steiner, Sutskever, Talwar, Tucker, Vanhoucke, Vasudevan,
  Vi\'{e}gas, Vinyals, Warden, Wattenberg, Wicke, Yu, and
  Zheng}]{Tensorflow.2015}
Abadi M, Agarwal A, Barham P, Brevdo E, Chen Z, Citro C, Corrado GS, Davis A,
  Dean J, Devin M, Ghemawat S, Goodfellow I, Harp A, Irving G, Isard M, Jia Y,
  Jozefowicz R, Kaiser L, Kudlur M, Levenberg J, Man\'{e} D, Monga R, Moore S,
  Murray D, Olah C, Schuster M, Shlens J, Steiner B, Sutskever I, Talwar K,
  Tucker P, Vanhoucke V, Vasudevan V, Vi\'{e}gas F, Vinyals O, Warden P,
  Wattenberg M, Wicke M, Yu Y, Zheng X (2015) {TensorFlow}: Large-scale machine
  learning on heterogeneous systems.
  \urlprefix\url{https://www.tensorflow.org/}, software available from
  tensorflow.org

\bibitem[{Al-Jebrni et~al.(2018)Al-Jebrni, Cai, and Jiang}]{AlJebrni.2018}
Al-Jebrni A, Cai H, Jiang L (2018) Predicting the next process event using
  convolutional neural networks. In: Wang Y, Sun Y, Wu X (eds) Proc. 2018 IEEE
  Int. Conf. Prog. INFORMATICS Comput, IEEE, 345 E 47TH ST, NEW YORK, NY 10017
  USA, Proceedings of the IEEE International Conference on Progress in
  Informatics and Computing, pp 332--338

\bibitem[{Augusto et~al.(2018)Augusto, Conforti, Dumas, La~Rosa, Maggi,
  Marrella, Mecella, and Soo}]{Augusto.2018}
Augusto A, Conforti R, Dumas M, La~Rosa M, Maggi FM, Marrella A, Mecella M, Soo
  A (2018) Automated discovery of process models from event logs: Review and
  benchmark. IEEE Transactions on Knowledge and Data Engineering 31(4):686--705

\bibitem[{Bandis et~al.(2018)Bandis, Petridis, and Kapetanakis}]{Bandis.2018}
Bandis E, Petridis M, Kapetanakis S (2018) Business process workflow mining
  using machine learning techniques for the rail transport industry. In: Lect.
  Notes Comput. Sci. (including Subser. Lect. Notes Artif. Intell. Lect. Notes
  Bioinformatics), {Springer Verlag}, vol 11311 LNAI, pp 446--451,
  \doi{10.1007/978-3-030-04191-5_37}

\bibitem[{Camargo et~al.(2019)Camargo, Dumas, and
  Gonz{\'a}lez-Rojas}]{Camargo.2019}
Camargo M, Dumas M, Gonz{\'a}lez-Rojas O (2019) Learning accurate lstm models
  of business processes. In: Lect. Notes Comput. Sci. (including Subser. Lect.
  Notes Artif. Intell. Lect. Notes Bioinformatics), {Springer Verlag}, vol
  11675 LNCS, pp 286--302, \doi{10.1007/978-3-030-26619-6_19}

\bibitem[{{Di Francescomarino} et~al.(2017){Di Francescomarino}, Ghidini,
  {Maria Maggi}, Petrucci, and Yeshchenko}]{DiFrancescomarino.2017}
{Di Francescomarino} C, Ghidini C, {Maria Maggi} F, Petrucci G, Yeshchenko A
  (2017) An eye into the future: Leveraging a-priori knowledge in predictive
  business process monitoring. In: International Conference on Business Process
  Management, SPRINGER, pp 252----268

\bibitem[{{Di Francescomarino} et~al.(2018){Di Francescomarino}, Ghidini,
  Maggi, and Milani}]{DiFrancescomarino.2018}
{Di Francescomarino} C, Ghidini C, Maggi FM, Milani F (2018) Predictive process
  monitoring methods: Which one suits me best? In: International Conference on
  Business Process Management, pp 462--479

\bibitem[{{Di Mauro} et~al.(2019){Di Mauro}, Appice, and Basile}]{DiMauro.2019}
{Di Mauro} N, Appice A, Basile TM (2019) Activity prediction of business
  process instances with inception cnn models. In: Lect. Notes Comput. Sci.
  (including Subser. Lect. Notes Artif. Intell. Lect. Notes Bioinformatics),
  {Springer Verlag}, vol 11946 LNAI, pp 348--361,
  \doi{10.1007/978-3-030-35166-3_25}

\bibitem[{Evermann et~al.(2016)Evermann, Rehse, and Fettke}]{Evermann.2016}
Evermann J, Rehse JR, Fettke P (2016) A deep learning approach for predicting
  process behaviour at runtime. In: International Conference on Business
  Process Management, SPRINGER, pp 327----338

\bibitem[{Evermann et~al.(2017)Evermann, Rehse, and Fettke}]{Evermann.2017}
Evermann J, Rehse JR, Fettke P (2017) Predicting process behaviour using deep
  learning. Decis Support Syst 100:129--140, \doi{10.1016/j.dss.2017.04.003}

\bibitem[{Ezpeleta et~al.(2018)Ezpeleta, Fabra, and Alvarez}]{Ezpeleta.2018}
Ezpeleta J, Fabra J, Alvarez P (2018) On the use of log-based model checking,
  clustering and machine learning for process behavior prediction. In: 2018
  FIFTH Int. Conf. Soc. NETWORKS Anal. Manag. Secur, IEEE, 345 E 47TH ST, NEW
  YORK, NY 10017 USA, pp 209--214

\bibitem[{Harane and Rathi(2020)}]{Harane.2020}
Harane N, Rathi S (2020) Comprehensive survey on deep learning approaches in
  predictive business process monitoring. In: Gunjan VK, Zurada JM, Raman B,
  Gangadharan GR (eds) Modern Approaches in Machine Learning and Cognitive
  Science: A Walkthrough, Studies in Computational Intelligence, vol 885,
  {Springer International Publishing}, Cham, pp 115--128,
  \doi{10.1007/978-3-030-38445-6_9}

\bibitem[{Hinkka et~al.(2019)Hinkka, Lehto, Heljanko, and Jung}]{Hinkka.2019}
Hinkka M, Lehto T, Heljanko K, Jung A (2019) Classifying process instances
  using recurrent neural networks. Lecture Notes in Business Information
  Processing 342:313--324, \doi{10.1007/978-3-030-11641-5_25}

\bibitem[{Khan et~al.(2018)Khan, Le, Do, Tran, Ghose, Dam, and
  Sindhgatta}]{Khan.2018}
Khan A, Le H, Do K, Tran T, Ghose A, Dam H, Sindhgatta R (2018)
  Memory-augmented neural networks for predictive process analytics. arXiv
  preprint arXiv:180200938

\bibitem[{Kitchenham et~al.(2009)Kitchenham, Brereton, Budgen, Turner, Bailey,
  and Linkman}]{Kitchenham.2009}
Kitchenham B, Brereton OP, Budgen D, Turner M, Bailey J, Linkman S (2009)
  Systematic literature reviews in software engineering--a systematic
  literature review. Information and software technology 51(1):7--15

\bibitem[{Kratsch et~al.(2020)Kratsch, Manderscheid, R{\"o}glinger, and
  Seyfried}]{Kratsch.2020}
Kratsch W, Manderscheid J, R{\"o}glinger M, Seyfried J (2020) Machine learning
  in business process monitoring: A comparison of deep learning and classical
  approaches used for outcome prediction. Bus Inf Syst Eng
  \doi{10.1007/s12599-020-00645-0}

\bibitem[{LeCun et~al.(2015)LeCun, Bengio, and Hinton}]{LeCun.2015}
LeCun Y, Bengio Y, Hinton G (2015) Deep learning. Nature 521(7553):436--444,
  \doi{10.1038/nature14539}

\bibitem[{Lin et~al.(2019)Lin, Wen, and Wang}]{Lin.2019}
Lin L, Wen L, Wang J (2019) Mm-pred: A deep predictive model for
  multi-attribute event sequence. In: Proceedings of the 2019 SIAM
  International Conference on Data Mining, pp 118----126

\bibitem[{Maggi et~al.(2014)Maggi, Di~Francescomarino, Dumas, and
  Ghidini}]{Maggi.2014}
Maggi FM, Di~Francescomarino C, Dumas M, Ghidini C (2014) Predictive monitoring
  of business processes. In: International conference on advanced information
  systems engineering, Springer, pp 457--472

\bibitem[{M{\'a}rquez-Chamorro et~al.(2017)M{\'a}rquez-Chamorro, Resinas, and
  Ruiz-Cortes}]{MarquezChamorro.2018}
M{\'a}rquez-Chamorro AE, Resinas M, Ruiz-Cortes A (2017) Predictive monitoring
  of business processes: a survey. IEEE Transactions on Services Computing
  11(6):962--977

\bibitem[{Mehdiyev et~al.(2017)Mehdiyev, Fettke, and Evermann}]{Mehdiyev.2017}
Mehdiyev N, Fettke P, Evermann J (2017) A multi-stage deep learning approach
  for business process event prediction. In: Loucopoulos P, Manolopoulos Y,
  Pastor O, Theodoulidis B, Zdravkovic J (eds) 2017 IEEE 19TH Conf. Bus.
  INFORMATICS (CBI), VOL 1, IEEE, 345 E 47TH ST, NEW YORK, NY 10017 USA,
  Conference on Business Informatics, vol~1, pp 119--128,
  \doi{10.1109/CBI.2017.46}

\bibitem[{Mehdiyev et~al.(2020)Mehdiyev, Evermann, and Fettke}]{Mehdiyev.2020}
Mehdiyev N, Evermann J, Fettke P (2020) A novel business process prediction
  model using a deep learning method. Bus Inf Syst Eng 62(2):143--157,
  \doi{10.1007/s12599-018-0551-3}

\bibitem[{Metzger and Bohn(2017)}]{Metzger.2017b}
Metzger A, Bohn P (2017) Risk-based proactive process adaptation. In:
  International Conference on Service-Oriented Computing, SPRINGER, pp
  351--366, \doi{10.1007/978-3-319-69035-3}

\bibitem[{Metzger and F{\"o}cker(2017)}]{Metzger.2017}
Metzger A, F{\"o}cker F (2017) Predictive business process monitoring
  considering reliability estimates. In: International Conference on Advanced
  Information Systems Engineering, SPRINGER, pp 445--460,
  \doi{10.1007/978-3-319-59536-8_28}

\bibitem[{Metzger and Neubauer(2018)}]{Metzger.2018}
Metzger A, Neubauer A (2018) Considering non-sequential control flows for
  process prediction with recurrent neural networks. In: Bures T, Angelis L
  (eds) 44TH EUROMICRO Conf. Softw. Eng. Adv. Appl. (SEAA 2018), IEEE, 345 E
  47TH ST, NEW YORK, NY 10017 USA, EUROMICRO Conference Proceedings, pp
  268--272, \doi{10.1109/SEAA.2018.00051}

\bibitem[{Metzger et~al.(2019)Metzger, Neubauer, Bohn, and Pohl}]{Metzger.2019}
Metzger A, Neubauer A, Bohn P, Pohl K (2019) Proactive process adaptation using
  deep learning ensembles. In: Giorgini P, Weber B (eds) Adv. Inf. Syst. Eng.
  (CAISE 2019), {SPRINGER INTERNATIONAL PUBLISHING AG}, GEWERBESTRASSE 11,
  CHAM, CH-6330, SWITZERLAND, Lecture Notes in Computer Science, vol 11483, pp
  547--562, \doi{10.1007/978-3-030-21290-2_34}

\bibitem[{Navarin et~al.(2017)Navarin, Vincenzi, Polato, and
  Sperduti}]{Navarin.2017}
Navarin N, Vincenzi B, Polato M, Sperduti A (2017) Lstm networks for data-aware
  remaining time prediction of business process instances. In: 2017 IEEE Symp.
  Ser. Comput. Intell, IEEE, 345 E 47TH ST, NEW YORK, NY 10017 USA, pp
  3474--3480

\bibitem[{Nolle et~al.(2016)Nolle, Seeliger, and
  M{\"u}hlh{\"a}user}]{Nolle.2016}
Nolle T, Seeliger A, M{\"u}hlh{\"a}user M (2016) Unsupervised anomaly detection
  in noisy business process event logs using denoising autoencoders. In: Lect.
  Notes Comput. Sci. (including Subser. Lect. Notes Artif. Intell. Lect. Notes
  Bioinformatics), {Springer Verlag}, vol 9956 LNAI, pp 442--456,
  \doi{10.1007/978-3-319-46307-0_28}

\bibitem[{Nolle et~al.(2018)Nolle, Seeliger, and
  M{\"u}hlh{\"a}user}]{Nolle.2018}
Nolle T, Seeliger A, M{\"u}hlh{\"a}user M (2018) Binet: Multivariate business
  process anomaly detection using deep learning. In: Lect. Notes Comput. Sci.
  (including Subser. Lect. Notes Artif. Intell. Lect. Notes Bioinformatics),
  {Springer Verlag}, vol 11080 LNCS, pp 271--287,
  \doi{10.1007/978-3-319-98648-7_16}

\bibitem[{Park and Song(2020)}]{Park.2020}
Park G, Song M (2020) Predicting performances in business processes using deep
  neural networks. Decis Support Syst 129, \doi{10.1016/j.dss.2019.113191}

\bibitem[{Pasquadibisceglie et~al.(2019)Pasquadibisceglie, Appice, Castellano,
  and Malerba}]{Pasquadibisceglie.2019}
Pasquadibisceglie V, Appice A, Castellano G, Malerba D (2019) Using
  convolutional neural networks for predictive process analytics. In: 2019 Int.
  Conf. Process Min. (ICPM 2019), IEEE, 345 E 47TH ST, NEW YORK, NY 10017 USA,
  pp 129--136, \doi{10.1109/ICPM.2019.00028}

\bibitem[{Schoenig et~al.(2018)Schoenig, Jasinski, Ackermann, and
  Jablonski}]{Schoenig.2018}
Schoenig S, Jasinski R, Ackermann L, Jablonski S (2018) Deep learning process
  prediction with discrete and continuous data features. In: Damiani E,
  Spanoudakis G, Maciaszek L (eds) Proc. 13TH Int. Conf. Eval. Nov. APPROACHES
  TO Softw. Eng, SCITEPRESS, AV D MANUELL, 27A 2 ESQ, SETUBAL, 2910-595,
  PORTUGAL, pp 314--319, \doi{10.5220/0006772003140319}

\bibitem[{Tax et~al.(2017)Tax, Verenich, {La Rosa}, and Dumas}]{Tax.2017}
Tax N, Verenich I, {La Rosa} M, Dumas M (2017) Predictive business process
  monitoring with lstm neural networks. In: Lect. Notes Comput. Sci. (including
  Subser. Lect. Notes Artif. Intell. Lect. Notes Bioinformatics), {Springer
  Verlag}, vol 10253 LNCS, pp 477--492, \doi{10.1007/978-3-319-59536-8_30}

\bibitem[{Tax et~al.(2018)Tax, Teinemaa, and van Zelst}]{Tax.2018}
Tax N, Teinemaa I, van Zelst SJ (2018) An interdisciplinary comparison of
  sequence modeling methods for next-element prediction. arXiv preprint
  arXiv:181100062

\bibitem[{Taymouri et~al.(2020)Taymouri, {La Rosa}, Erfani, Bozorgi, and
  Verenich}]{Taymouri.2020}
Taymouri F, {La Rosa} M, Erfani S, Bozorgi ZD, Verenich I (2020) Predictive
  business process monitoring via generative adversarial nets: The case of next
  event prediction. \urlprefix\url{http://arxiv.org/pdf/2003.11268v2}

\bibitem[{Teinemaa et~al.(2018)Teinemaa, Dumas, Leontjeva, and
  Maggi}]{Teinemaa.2018}
Teinemaa I, Dumas M, Leontjeva A, Maggi FM (2018) Temporal stability in
  predictive process monitoring. DATA Min Knowl Discov 32(5, SI):1306--1338,
  \doi{10.1007/s10618-018-0575-9}

\bibitem[{Teinemaa et~al.(2019)Teinemaa, Dumas, {La Rosa}, and
  Maggi}]{Teinemaa.2019}
Teinemaa I, Dumas M, {La Rosa} M, Maggi FM (2019) Outcome-oriented predictive
  process monitoring: Review and benchmark. ACM Transactions on Knowledge
  Discovery from Data (TKDD) 13(2):1--57

\bibitem[{Theis and Darabi(2019)}]{Theis.2019}
Theis J, Darabi H (2019) Decay replay mining to predict next process events.
  IEEE ACCESS 7:119787--119803, \doi{10.1109/ACCESS.2019.2937085}

\bibitem[{Verenich et~al.(2019)Verenich, Dumas, Rosa, Maggi, and
  Teinemaa}]{Verenich.2019}
Verenich I, Dumas M, Rosa ML, Maggi FM, Teinemaa I (2019) Survey and
  cross-benchmark comparison of remaining time prediction methods in business
  process monitoring. ACM Transactions on Intelligent Systems and Technology
  (TIST) 10(4):1--34

\bibitem[{Wahid et~al.(2019)Wahid, Adi, Bae, and Choi}]{Wahid.2019}
Wahid NA, Adi TN, Bae H, Choi Y (2019) Predictive business process monitoring
  -- remaining time prediction using deep neural network with entity embedding.
  Procedia Comput Sci 161:1080--1088, \doi{10.1016/j.procs.2019.11.219}

\bibitem[{Wang et~al.(2019)Wang, Yu, Liu, and Sun}]{Wang.2019}
Wang J, Yu D, Liu C, Sun X (2019) Outcome-oriented predictive process
  monitoring with attention-based bidirectional lstm neural networks. In:
  Bertino E, Chang CK, Chen P, Damiani E, Goul M, Oyama K (eds) 2019 IEEE Int.
  Conf. WEB Serv. (IEEE ICWS 2019), IEEE, 345 E 47TH ST, NEW YORK, NY 10017
  USA, pp 360--367, \doi{10.1109/ICWS.2019.00065}

\bibitem[{{XES Working Group}(2016)}]{XES.2016d}
{XES Working Group} (2016) Ieee standard for extensible event stream (xes) for
  achieving interoperability in event logs and event streams. IEEE Std 1849 pp
  1--50

\bibitem[{Zhang and Yang(2017)}]{Zhang.2017}
Zhang Y, Yang Q (2017) {A Survey on Multi-Task Learning}. arXiv:180200938 pp
  1--20, \urlprefix\url{http://arxiv.org/abs/1707.08114}, \eprint{1707.08114}

\end{thebibliography}
\end{document}